\pdfoutput=1

\documentclass[letterpaper, 10pt, conference]{ieeeconf}      
 
\IEEEoverridecommandlockouts                              

\usepackage{multicol}
\usepackage{multirow}
\usepackage{algpseudocode}
\usepackage{color}
\usepackage{graphicx}
\usepackage{bm}
\usepackage{amsmath}
\usepackage{amssymb}
\usepackage{tabularx}
\usepackage{subcaption}
\usepackage{url}
\usepackage{cite}
\usepackage{prettyref}
\usepackage{booktabs}
\usepackage{siunitx}
\usepackage{comment}
\usepackage{booktabs}
\usepackage{hyperref}
\usepackage{tabularx, booktabs}
\usepackage[capitalise]{cleveref}

\usepackage{makecell} 
\usepackage[table]{xcolor} 
\usepackage{pifont} 
\usepackage{tabularx} 
\definecolor{Gray}{gray}{0.9}

\usepackage[all]{hypcap}
\usepackage{comment}

\usepackage[ruled]{algorithm2e} 

\SetAlFnt{\small}
\SetAlCapFnt{\small}
\SetAlCapNameFnt{\small}
\SetAlCapHSkip{0pt}

\newrefformat{fig}{Figure~\ref{#1}}
\newrefformat{sec}{Section~\ref{#1}}
\newrefformat{tab}{Table~\ref{#1}}
\newrefformat{alg}{Algorithm~\ref{#1}}
\newrefformat{eqn}{Equation~\ref{#1}}

\hypersetup{
    colorlinks,
    linkcolor={blue!80!black},
    citecolor={blue!80!black},
    urlcolor={blue!100!black}
}

\makeatletter
\newcommand{\printfnsymbol}[1]{%
  \textsuperscript{\@fnsymbol{#1}}%
}
\makeatother
\newcommand{\sysname}{WeatherGS}

\title{\LARGE \bf \sysname: 3D Scene Gaussian Splatting for Snow Removal Rendering}
\title{\LARGE \bf \sysname: 3D Scene Reconstruction in Adverse Weather Conditions via Gaussian Splatting }

\author{ Chenghao Qian~$^{*}$, Yuhu Guo~$^{*}$, Wenjing Li, Gustav Markkula
\small{
\thanks{C. Qian, W. Li and G. Markkula are with the Institute for Transport Studies, the University of Leeds}
\thanks{Y. Guo is with the Electrical and Computer Engineering, Carnegie Mellon University, USA}
\thanks{$^{*}$~The first two author are equally contribution}
}
}

\begin{document}
\maketitle

\begin{abstract}

3D Gaussian Splatting (3DGS) has gained significant attention for 3D scene reconstruction, but still suffers from complex outdoor environments, especially under adverse weather. This is because 3DGS treats the artifacts caused by adverse weather as part of the scene and will directly reconstruct them, largely reducing the clarity of the reconstructed scene. To address this challenge, we propose WeatherGS, a 3DGS-based framework for reconstructing clear scenes from multi-view images under different weather conditions. Specifically, we explicitly categorize the multi-weather artifacts into the dense particles and lens occlusions that have very different characters, in which the former are caused by snowflakes and raindrops in the air, and the latter are raised by the precipitation on the camera lens. In light of this, we propose a \textit{dense-to-sparse} preprocess strategy, which sequentially removes the dense particles by an Atmospheric Effect Filter (AEF) and then extracts the relatively sparse occlusion masks with a Lens Effect Detector (LED). Finally, we train a set of 3D Gaussians by the processed images and generated masks for excluding occluded areas, and accurately recover the underlying clear scene by Gaussian splatting. We conduct a diverse and challenging benchmark to facilitate the evaluation of 3D reconstruction under complex weather scenarios. Extensive experiments on this benchmark demonstrate that our WeatherGS consistently produces high-quality, clean scenes across various weather scenarios, outperforming existing state-of-the-art methods. See project:\href{https://jumponthemoon.github.io/weather-gs}{https://jumponthemoon.github.io/weather-gs}

\end{abstract}

\section{Introduction}


3D scene reconstruction has significant applications in diverse fields, including robotics, virtual reality (VR), and autonomous driving. 
However, the acquisition of accurate scene representations under adverse weather conditions presents significant challenges. 
Specifically, the presence of weather particles such as snowflakes and raindrops can drastically degrade captured image quality and hinder precise reconstruction. In more extreme cases, precipitation can adhere to imaging sensors that cause significant occlusion and distortion,  making the reconstruction even more difficult.

\begin{figure}[t]
  \centering
    \centering
    \includegraphics[width=\linewidth]{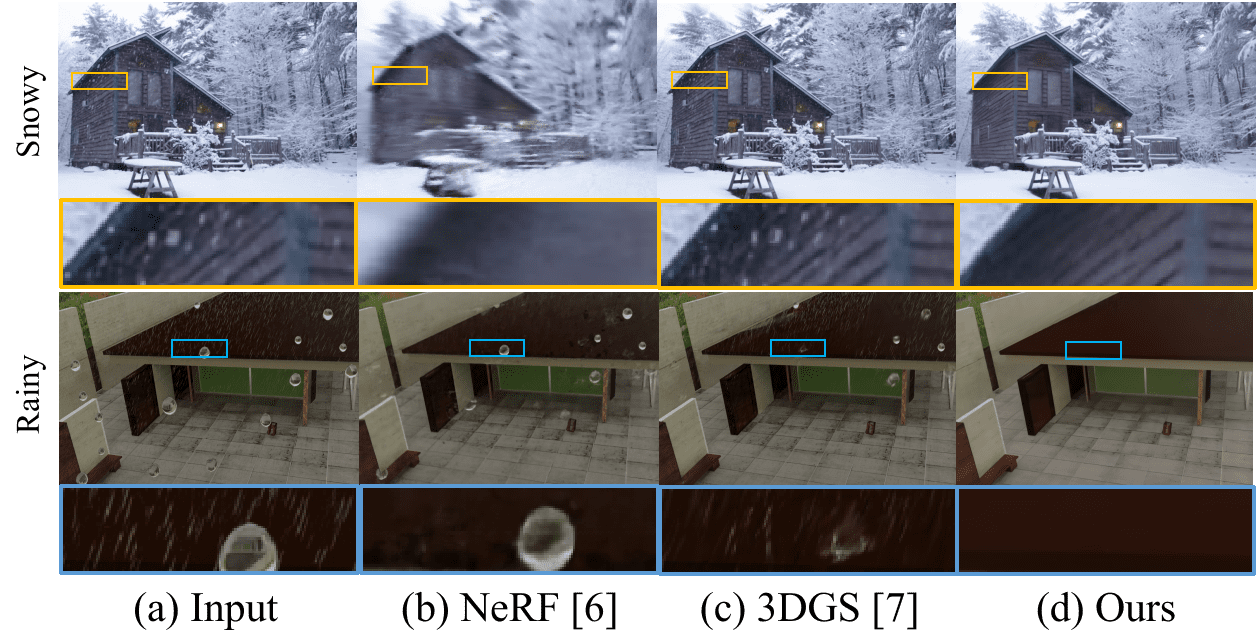}
    \caption{Rendering examples under adverse weather conditions. NeRF introduces blur effects and lens occlusions, while 3DGS reconstructs scenes with dense weather particles and similar occlusions, significantly obscuring visibility. In contrast, the proposed WeatherGS effectively removes these artifacts, reconstructing clean 3D scenes and rendering artifact-free images.}
    \vspace{-20pt}

  \label{fig:intro1}
\end{figure}

Existing methods primarily address challenges related to low illumination \cite{wang2023lighting,cui_aleth_nerf}  and blur effects \cite{Lee_2023_CVPR,chen2024deblur}, neglecting the impact of weather-related issues. 
Recently, DerainNeRF \cite{li2024derainnerf} proposes a Neural Radiance Field (NeRF)--based framework \cite{mildenhall2021nerf} that can effectively remove water droplets from camera lenses.
However, the continuous volumetric representation within NeRF 
makes it susceptible to inaccuracies when dealing with dynamic weather particles like falling snowflakes and raindrops.
This is because the varying positions, shapes, and visibility of these particles across different views introduce inconsistencies during reconstruction training, resulting in blurred or inaccurately rendered regions, as shown in \cref{fig:intro1}b. 
Additionally, the high computational cost of NeRF limits the applicability of NeRF-based methods in real-time applications.

%

More recently, 3D Gaussian splatting (3DGS) \cite{kerbl3Dgaussians} has emerged as a promising alternative to NeRF due to its high visual fidelity and computational efficiency. 3DGS represents each 3D point as a flexible Gaussian splat, making it highly adaptable to scene dynamics, such as falling weather particles. The Gaussian distribution inherently filters and smooths small-scale noise, offering the potential to reduce weather-related artifacts. However, dense particles from raindrops and snowflakes, as well as significant occlusions caused by precipitation on the lens, are comparatively consistent across multiple views and can therefore be directly reconstructed by 3DGS, which substantially reduces the clarity of the scene, as shown in \cref{fig:intro1}c. This issue has not been extensively explored, thus hindering the application of 3DGS under challenging weather conditions.






To address these limitations, we propose WeatherGS, a novel 3DGS-based framework capable of reconstructing clean 3D scenes from complex weather scenarios. 
WeatherGS adopts a \textit{dense-to-sparse} strategy to preprocess multi-view images by first removing weather particles with an Atmospheric Effect Filter (AEF), followed by extracting occlusions with a Lens Effect Detector (LED).
This step generates a cleaner set of images, providing a robust foundation for 3D reconstruction training. 
%
Finally, we use the pre-processed images and generated masks to train a set of 3D Gaussians through Gaussian splatting, enabling the reconstruction of a clean scene.
To summarize, our contributions are three-fold: (1) We provide a comprehensive benchmark to evaluate the effectiveness of 3D reconstruction methods under adverse weather conditions. The benchmark includes both synthetic and real-world scenarios, covering snowy and rainy weather conditions.
(2) To address weather-induced dense artifacts and lens occlusions during 3D reconstruction, we propose a novel 3DGS-based framework, WeatherGS, capable of reconstructing clean 3D scenes from multi-weather scenarios.
(3) The qualitative and quantitative experiment results show that WeatherGS can effectively handle weather-induced dense particles, occlusions and distortions, enabling high-quality 3D scene reconstruction and real-time rendering across diverse weather conditions.

\label{sec:intro}

\section{Related Work}
\subsection{3D Gaussian Splatting}

3D Gaussian Splatting (3DGS) is an emerging paradigm within the field of radiance field representation. It explicitly encodes the 3D scene using learnable 3D Gaussians as a flexible and efficient representation, with each Gaussian representing a small volume with associated radiance properties. These Gaussians are then optimized by iteratively refining Gaussian parameters under the supervision of multi-view images to accurately model the scene. By combining the efficiency of Gaussian primitives with advanced optimization techniques, 3DGS achieves remarkable real-time rendering performance, surpassing NeRF in terms of both visual quality and computational performance,
making it effective for handling complex scenes and producing high quality outputs.
The versatility of 3DGS enables its application across a wide range of areas, including dynamic scene reconstruction \cite{guo2024motion,luiten2023dynamic,yang2023real,chen2023periodic,zhang2024gaussian}, content creation \cite{ren2023dreamgaussian4d, voleti2024sv3d ,tang2024lgm} and scene editing \cite{chen2024gaussianeditor,liu2024stylegaussian,ye2023gaussian}. Despite these advancements, the application of 3DGS in the context of challenging weather conditions remains unexplored.


%

\subsection{Weather Effect Removal}
Most existing weather effect removal techniques focus on removing weather-related visual artifacts in 2D spaces. Early methods for weather artifact removal utilized traditional image processing and mathematical techniques to model physical priors based on empirical observations \cite{roth2005fields,rudin1992nonlinear}. Later, Convolutional Neural Networks (CNNs) were extensively explored for tasks such as desnowing\cite{JSTASRChen,zhang2021deep} and deraining \cite{ren2019progressive,qian2018attentive,wen2023video}. Desnow-Net \cite{liu2018desnownet} was one of the first CNN-based methods proposed to remove snow from images. Subsequently, DDMSNet \cite{zhang2021deep} introduced a deep dense multiscale network that leverages semantic and geometric priors for snow removal. Deraining methods focus on removing rain streaks. For example, Li et al. \cite{li2018recurrent} used a recurrent network to decompose rain layers into different layers of various streak types, facilitating rain removal. To remove raindrops adhered to a glass window or camera lens, Qian et al. \cite{qian2018attentive} propose an attention mechanism within a generative network using adversarial training, allowing the network to pay more attention to the raindrop regions for removal. More recently, some works have proposed unified architectures \cite{qian2025allweather,valanarasu2021transweather,li2020all} to address these tasks within a single network but they are still limited to 2D spaces. 
In the realm of 3D space weather effect removal, DerainNeRF \cite{li2024derainnerf} pioneered a NeRF-based approach to address image degradation caused by waterdrops. However, it exhibits limitations in handling diverse real-world scenarios, such as those involving snowflakes and rain streaks, or more complex scenes with hybrid effects like rain streaks combined with occluded camera lenses.

\begin{figure*}[ht]
    \centering
    \includegraphics[width=\linewidth]{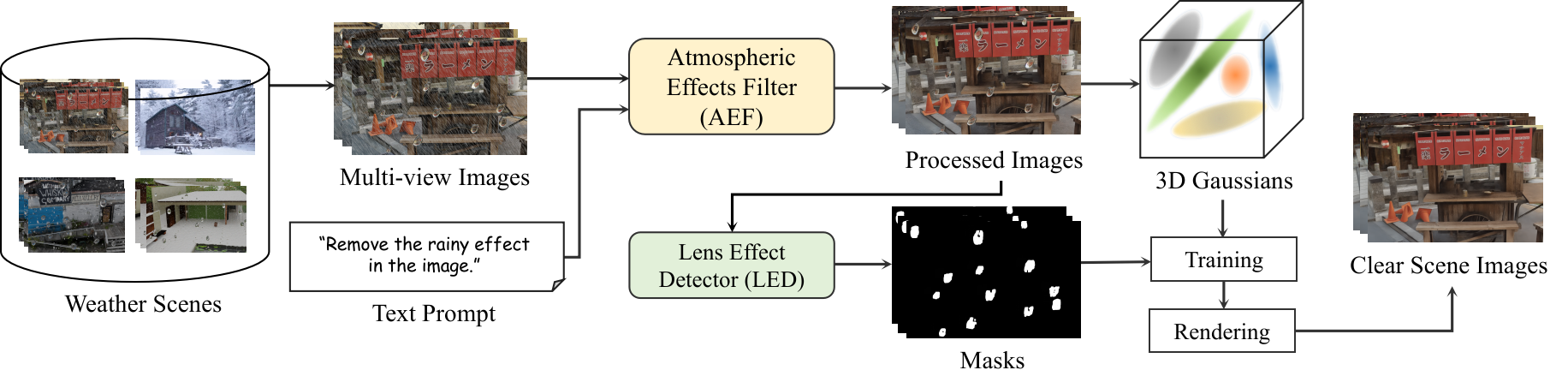}
    \caption{The overview of WeatherGS. The WeatherGS preprocesses multi-view images by removing weather-related visual artifacts. An atmospheric filter removes dense particles like raindrops and snowflakes, while a lens effect detector identifies and masks occlusions caused by precipitation on the camera lens. To ensure high-quality 3D scene reconstruction, we train the preprocessed images with 3D Gaussian splatting excluding the occlusion area to model the clear scene's geometry and appearance.}
    \label{fig:pipeline}
    \vspace{-4mm}
\end{figure*}
\section{Method}

\subsection{Prelimineries}
\label{sec:pre}
3DGS represents a scene by defining an explicit radiance field to model light distribution in a 3D space. It initializes the scene with a collection of 3D Gaussians, each parameterized by its center position $\mu$, opacity $\alpha$, 3D covariance matrix $\Sigma$, and  RGB color (or spherical harmonics) $c$. These 3D Gaussians are then projected into the 2D image space for rendering. Given the viewing transformation $W$ and 3D covariance matrix $\Sigma$, the projected 2D covariance matrix $\Sigma'$ is computed using:

\begin{equation}
\Sigma' = JW\Sigma W^{\top} J^{\top},
\end{equation}
where $J$ is the Jacobian of the affine approximation of the projective transformation. Gaussian distances are calculated through the viewing transformation $W$ and sorted by proximity. Alpha compositing is then used to determine the final pixel color:
\begin{equation}
C= \sum_{n=1}^{|N|} c_n \alpha'_n \prod_{j=1}^{n-1} (1 - \alpha'_j),
\end{equation}
where $c_n$ denotes the learned color, while $N$ is the number of Gaussian kernels. The final opacity is determined by the product of learned opacity $\alpha_n$ and the Gaussian, described as follows:
\begin{equation}
    G(x') = \alpha_n \exp  \left(-\frac{1}{2}(x' - \mu'_n)^{\top} {{\Sigma}^{'}_n}^{-1} (x' - \mu'_n)
   \right),
\end{equation}
where $x'$ represents the point in a 3D space and $\mu'_n$ denotes the mean of 3D Gaussian  parametrized position. The rendered pixels of 3D points are then compared with the original input image pixels to compute photometric loss to update 3D Gaussian parameters.
\subsection{Multi-view Image Preprocessing}
When capturing outdoor environments with cameras, images inevitably contain dense artifacts from weather conditions like snowflakes and rain streaks. Worse, these may adhere to the lens, causing significant occlusions.
Such artifacts can mislead the 3D reconstruction model, as they are often interpreted as part of the scene and directly reconstructed, resulting in texture inconsistencies and distorted 3D models.
%

To address these challenges, we propose a novel framework that effectively generates high-quality multi-view images by removing multi-weather artifacts, thereby helping the 3DGS model render clean scenes, as shown in \cref{fig:pipeline}.
To be specific, we classify the weather artifacts into two categories: the weather particles caused by the raindrops and snowflakes, and the occlusions caused by the precipitation on the lens.
We find that weather particles are generally dense but small in scale, while occlusions are relatively sparse but typically larger in size.
Therefore, we believe it is more reasonable to use different mechanisms to handle them separately.
Inspired by this,
we propose a \textit{dense-to-sparse} framework to process these two types of artifacts separately. First, the Atmospheric Effect Filter (AEF) is used to remove weather particles, followed by the application of the Lens Effect Detector (LED) to extract occlusion information.
Note that the AEF module assists the LED module in practice, as the LED can more easily extract occlusions once the dense particles have been removed.


\noindent \textbf{Atmospheric Effect Filter (AEF).} 
To remove weather particles,
we use diffusion models~\cite{rombach2021highresolution}, which are highly effective at reconstructing clean images from noisy inputs. 
However, directly applying diffusion models can lead to content distortion due to the inherent randomness in the diffusion process. 
Accordingly, we integrate weather-specific priors following Liu et al. \cite{liu2024diff} to guide the diffusion model, enabling it to selectively remove weather-induced artifacts while preserving the underlying scene details.
\begin{figure}[ht]
    \centering
    \includegraphics[width=1.06\linewidth]{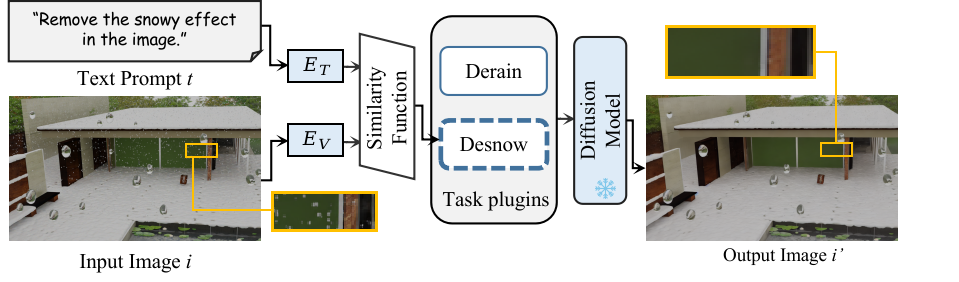}
    \caption{The pipeline of atmospheric effect filter. A snowy scene is processed using the text prompt ``Remove the snowy effect in the image." The image passes through encoder layers, and the system selects the ``Desnow" task via similarity functions to remove snowflakes. }
    \label{fig:method1}
    \vspace{-4mm}
\end{figure}

In detail, we extract semantic features from input weather images using a pre-trained CLIP \cite{radford2021learning} vision encoder as task guidance. 
These features are then utilized within ResNet \cite{he2016deep} and cross-attention layers in the diffusion model, along with the original images encoded by VAE \cite{kingma2013auto}, to train the model for a specific image restoration task. 
After training, the learned layers can act as task plugin and be integrated into diffusion models to guide the diffusion model in removing weather particles while preserving fine-grained spatial details. 
To facilitate the use of multiple plugins for different weather conditions, we implement a method that allows users to use text instructions and input images to automatically select the most appropriate plugins.
Specifically, we encode the input weather-removal text instruction \textit{t} and the input weather image \textit{i} using the encoders \(\text{E}_{\textit{T}}\) and \(\text{E}_{\textit{V}}\) into vectors, as introduced in \cite{liu2024diff}. 
We then compare the cosine similarity score between encoded features and apply a threshold \(\theta\) to select the best-fitting task plugin 
\(\mathcal{D^*}\) as follows:
\begin{equation}
\mathcal{D^*} = \begin{cases}
\mathcal{D}_{\text{derain}}, & \text{if } \cos(\text{E}_{\textit{T}}(t), \text{E}_{\textit{V}}(i)) > \theta \\
\mathcal{D}_{\text{desnow}}, & \text{otherwise}
\end{cases}
\end{equation}
With the selected task plugin's guidance, we can instruct the diffusion model to obtain a relatively clean weather image \textit{i'} with dense weather particles removed as illustrated in \cref{fig:method1}.

\noindent\textbf{Lens Effect Detector (LED).} Although AEF can remove most of weather particles in the image, 
it doesn't resolve the issue of occlusion caused by precipitation on the camera lens.
Thus, we incorporate LED which is composed of the detection module in AttGAN \cite{qian2018attentive} to identify the occluded areas, similar to \cite{li2024derainnerf}. 
Specifically, we feed the images processed by AEF into the LED to generate a confidence map \( C(x, y) \in [0, 1] \) that represents the likelihood of each pixel location \( (x, y) \) being occluded. 
For each image, a binary mask \( M \) is generated based on the following equation:
\begin{equation}
M(x, y) = \mathbb{I}(C(x, y) \geq t).
\end{equation}
Here, \( \mathbb{I} \) is the threshold function with the threshold \( t \). The function \( M(x, y) \) assigns a value of 1 to the pixel at location \( (x, y) \) if its confidence score exceeds the threshold \( t \), indicating it is covered by occlusions. Otherwise, the pixel is set to 0.

\subsection{3D Scene Reconstruction with Occlusion Masks}
With the restored images and generated lens effect masks, we reconstruct the scene using 3DGS. 
We initialize 3D Gaussians with a set of points from SfM \cite{schonberger2016structure}. The training and rendering follow steps introduced in \cref{sec:pre}. 
To update the parameters of 3D Gaussians, we utilize stochastic gradient descent with a combination of \( L_1 \) and D-SSIM loss functions. Additionally, masks generated by the lens effect detector are utilized to exclude occluded areas of the lens from loss computation during training. 
The \( L_1 \) loss is defined as: 
\begin{equation}
    \mathcal{L}_1 = \sum_{} \left\| \left( \hat{I}(\mathbf{t}) - I(\mathbf{t}) \right) \circ \left( \mathbf{1} - \mathbf{M} \right) \right\|^2,
\end{equation}
where \(\circ\) denotes element-wise multiplication, \(\hat{I}(\mathbf{t})\) represents the rendered pixel color at location \(\mathbf{t}\), \(I(\mathbf{t'})\) denotes the ground truth color value, and \(\mathbf{M}\) represents the lens effect mask.
Similarly, the D-SSIM loss is defined as:
\begin{equation}
    \mathcal{L}_{\text{D-SSIM}} = 1 - \text{SSIM}\left( \hat{I} \circ (\mathbf{1} - \mathbf{M}), I \circ (\mathbf{1} - \mathbf{M}) \right),
\end{equation}
where SSIM is the structural similarity index applied to the masked images. The final training loss is given by:
\begin{equation}
    \mathcal{L} = (1 - \lambda) \mathcal{L}_1 + \lambda \mathcal{L}_{\text{D-SSIM}},
\end{equation}
where \(\lambda \in [0, 1]\) is a weighting factor that balances the contribution of the two loss functions. After the training process is complete, we are able to recover the clear 3D scene. 





 \section{Experiments}
\begin{figure*}[ht]
    \centering
    \begin{subfigure}[b]{\linewidth}
        \centering
        \includegraphics[width=0.9\linewidth]{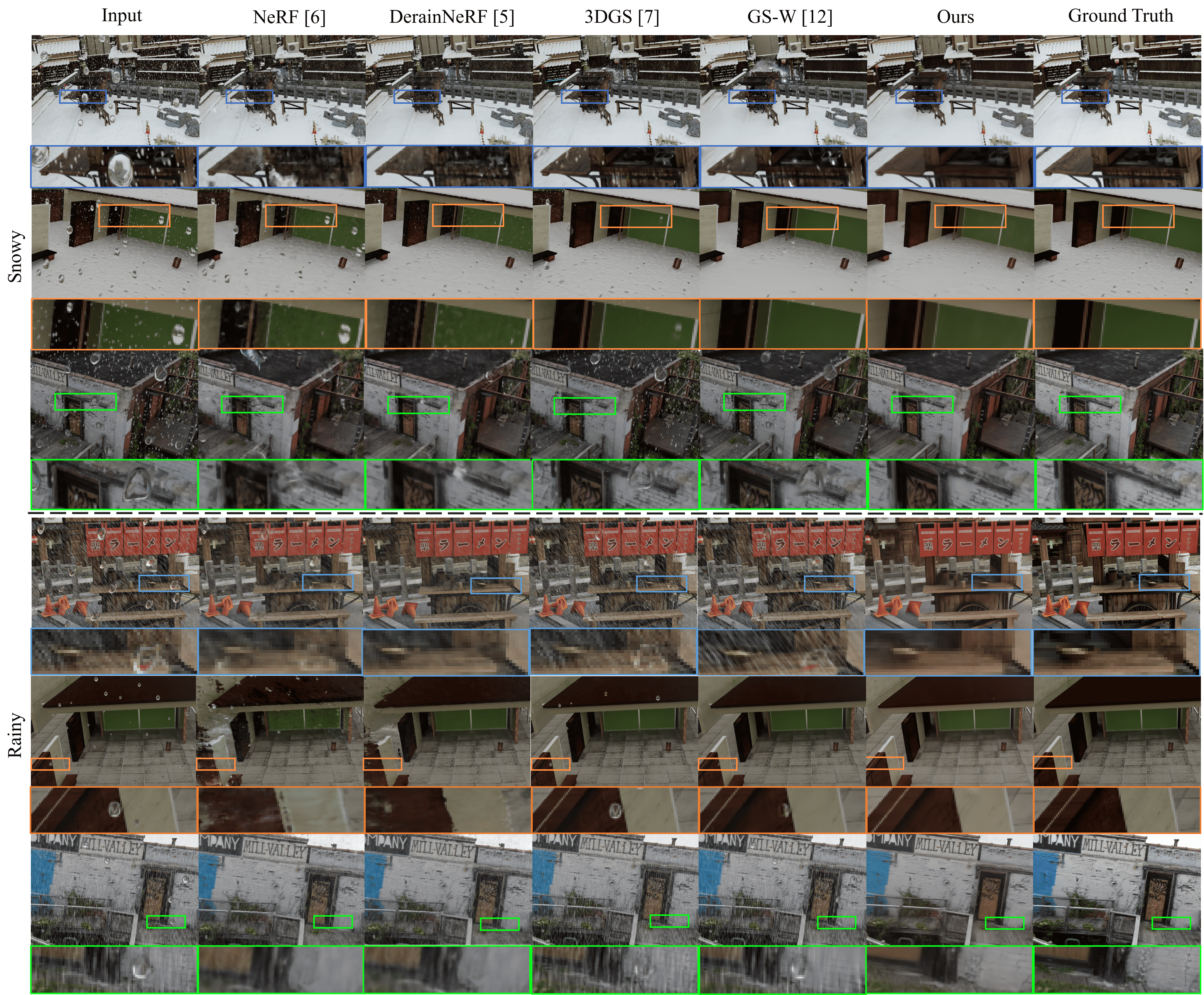}
        \footnotesize
        \caption{Synthetic}
        \label{fig:synthetic_results}
    \end{subfigure}
    \hfill
    \begin{subfigure}[b]{\linewidth}
        \centering
        \includegraphics[width=0.9\linewidth]{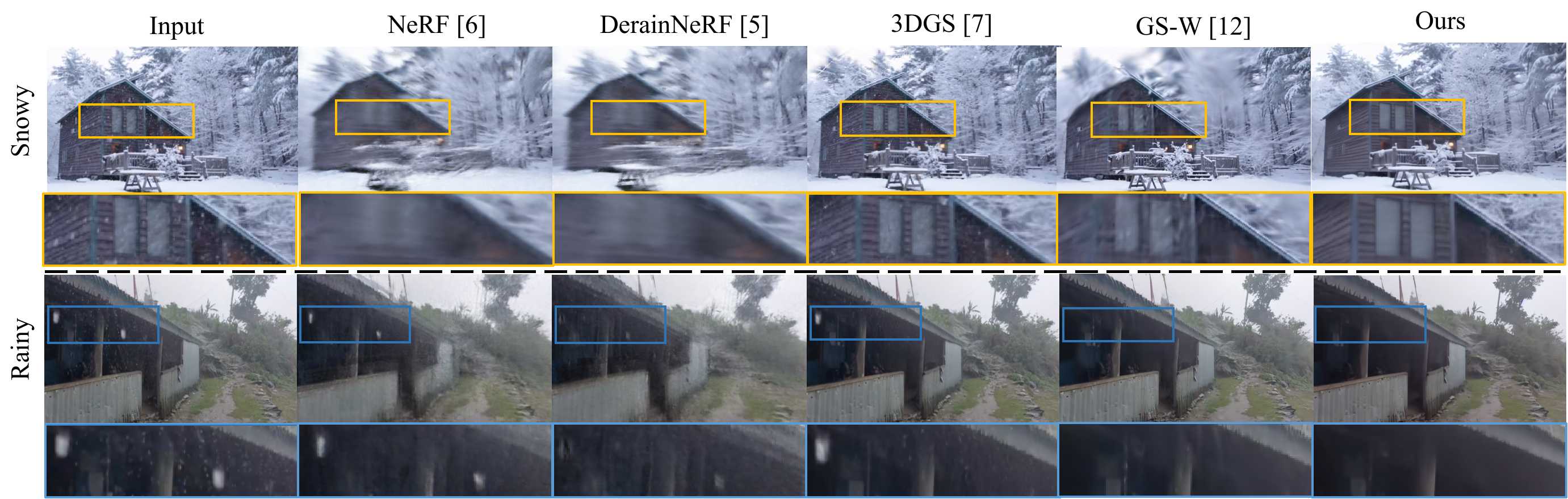}
        \footnotesize

        \caption{Real-world}
        \label{fig:real_results}
    \end{subfigure}

    \caption{Qualitative results on synthetic and real-world datasets under snowy and rainy conditions. NeRF-based methods show blurring and inaccurately rendered regions, while 3DGS-based approaches struggle to remove weather artifacts and camera occlusions. In contrast, WeatherGS successfully eliminates these weather-related disturbances, producing clearer, high-quality rendering results in both scenarios.}
    \label{fig:combined_results}
    \vspace{-5mm}
\end{figure*}

\begin{table*}[ht]
\centering
\caption{Comparison of PSNR, SSIM, and LPIPS for Different Models on Snowy Scenes}
\resizebox{\textwidth}{!}{
\begin{tabular}{lcccccccccccc}
\toprule
\multirow{2}{*}{\textbf{Model}} & \multicolumn{3}{c}{Tanabata-snow} & \multicolumn{3}{c}{Pool-snow} & \multicolumn{3}{c}{Factory-snow} & \multicolumn{3}{c}{Average} \\
\cmidrule(lr){2-4} \cmidrule(lr){5-7} \cmidrule(lr){8-10} \cmidrule(lr){11-13}
& PSNR↑ & SSIM↑ & LPIPS↓ & PSNR↑ & SSIM↑ & LPIPS↓ & PSNR↑ & SSIM↑ & LPIPS↓ & PSNR↑ & SSIM↑ & LPIPS↓ \\
\midrule
NeRF [6] & 21.377 & 0.729 & 0.215 & 21.820 & 0.731 & 0.376 & 21.666 & 0.612 & 0.383 & 21.621 & 0.691 & 0.325 \\
DerainNeRF [5] & 23.428 & 0.763 & 0.175 & 23.382 & 0.752 & 0.363 & 23.381 & 0.652 & 0.335 & 23.397 & 0.722 & 0.291 \\
3DGS [7] & 22.589 & \textbf{0.836} & 0.144 & 26.431 & 0.860 & 0.145 & 20.860 & \textbf{0.768} & 0.234 & 23.293 & \textbf{0.821} & 0.174 \\
GS-W [12] & 20.856 & 0.793 & 0.200 & 25.848 & 0.863 & 0.159 & 21.897 & 0.761 & 0.248 & 22.867 & 0.806 & 0.202 \\
\textbf{WeatherGS (Ours)} & \textbf{23.625} & 0.796 & \textbf{0.143} & \textbf{27.596} & \textbf{0.864} & \textbf{0.137} & \textbf{23.977} & 0.700 & \textbf{0.221} & \textbf{25.066} & 0.787 & \textbf{0.167} \\
\midrule
\end{tabular}}
\label{table: snowy}
\vspace{-2mm}
\end{table*}

\begin{table*}[h]
\centering
\caption{Comparison of PSNR, SSIM, and LPIPS for Different Models on Rainy Scenes}
\resizebox{\textwidth}{!}{
\begin{tabular}{lcccccccccccc}
\toprule
\multirow{2}{*}{\textbf{Model}} & \multicolumn{3}{c}{Tanabata-rain} & \multicolumn{3}{c}{Pool-rain} & \multicolumn{3}{c}{Factory-rain} & \multicolumn{3}{c}{Average} \\
\cmidrule(lr){2-4} \cmidrule(lr){5-7} \cmidrule(lr){8-10} \cmidrule(lr){11-13}
& PSNR↑ & SSIM↑ & LPIPS↓ & PSNR↑ & SSIM↑ & LPIPS↓ & PSNR↑ & SSIM↑ & LPIPS↓ & PSNR↑ & SSIM↑ & LPIPS↓ \\
\midrule
NeRF [6] & 20.860 & 0.564 & 0.370 & 25.324 & 0.771 & 0.183 & 14.305 & 0.276 & 0.543 & 20.163 & 0.537 & 0.365 \\
DerainNeRF [5] & 22.342 & 0.602 & 0.326 & 18.934 & 0.581 & 0.401 & 14.285 & 0.273 & 0.540 & 18.520 & 0.485 & 0.422 \\
3DGS [7] & 21.333 & 0.609 & 0.338 & \textbf{26.994} & \textbf{0.817} & 0.183 & 21.406 & 0.601 & 0.329 & 23.244 & 0.676 & 0.283 \\
GS-W [12] & 21.185 & 0.537 & 0.402 & 23.255 & 0.786 & 0.191 & 21.218 & 0.607 & 0.366 & 21.886 & 0.643 & 0.319 \\
\textbf{WeatherGS (Ours)} & \textbf{23.623} & \textbf{0.628} & \textbf{0.232} & 24.370 & 0.805 & \textbf{0.140} & \textbf{23.016} & \textbf{0.620} & \textbf{0.218} & \textbf{23.670} & \textbf{0.684} & \textbf{0.197} \\
\midrule
\end{tabular}}
\label{table: rainy}
\vspace{-4mm}
\end{table*}

 \subsection{Experimental Setup}
\noindent \textbf{Datasets.} 
To evaluate the effectiveness of the 3D reconstruction models under adverse weather conditions, we conduct a new, diverse, and challenging benchmark that includes both synthetic and real-world scenarios.
For the synthetic dataset, we use basic scenes provided by Deblur-NeRF \cite{li2022deblurnerf} and edit them in Blender \cite{Hess:2010:BFE:1893021} to generate multi-view weather images.
To simulate realistic snowy conditions, we employ a snow material add-on and apply snow cover to various scene elements. Additionally, we design a particle system capable of emitting particles with customized physics that closely resemble real snowflakes and raindrops. 
To enhance the authenticity of the captured images and better emulate real-world camera effects, we apply motion blur during the scene rendering process. 
Furthermore, to achieve a denser weather effect, we utilize OpenCV to enhance the overall visual realism of the weather conditions. In total, we created three distinct scenes (\textit{i.e.,} Tanabata, Factory, Pool), each featuring snowy and rainy weather. 
For the real-world dataset, we generate two scenes by extracting keyframes from publicly available online videos~\cite{videolink1, videolink2} captured in snowy and rainy weather, using them as multi-view image inputs.

\noindent \textbf{Implementation Details.} We develop the atmospheric effect filter using the pretrained task plugins from Liu et al.\cite{liu2024diff}. To remove atmospheric occlusions, we guide a pretrained Stable Diffusion model \cite{rombach2021highresolution} using text prompts based on specific weather effects. We train the Lens Effect Detector using the dataset provided by Qian et al. \cite{qian2018attentive} and further enhance the generated masks with averaging. Our 3D reconstruction model employs the 3D Gaussian Splatting framework \cite{kerbl3Dgaussians}, trained on multi-view images with corresponding masks for 30,000 iterations on a single RTX 3090 GPU. 



\noindent \textbf{Baselines and Evaluation Metric.}
To evaluate the visual quality of the reconstructed scenes, we display images rendered by our method against those produced by state-of-the-art 3D reconstruction models, including vanilla NeRF \cite{mildenhall2020nerf}, DerainNeRF \cite{li2024derainnerf}, vanilla 3DGS \cite{kerbl3Dgaussians} and GS-W \cite{zhang2024gaussian}. 
For quantitative evaluation, we benchmark our approach against these models using three standard metrics: Peak Signal-to-Noise Ratio (PSNR) , Structural Similarity Index Measure (SSIM), and Learned Perceptual Image Patch Similarity (LPIPS) \cite{zhang2018unreasonable}. 
Notably, for real-world scenes, we perform only qualitative comparisons as ground truth data is unavailable.

\subsection{Comparisons with the State-of-the-art Methods.}


\noindent{\textbf{Qualitative Analysis}.}
Fig.~\ref{fig:combined_results} shows the qualitative comparisons between the proposed WeatherGS and the baselines.
We can obtain several observations.
First, WeatherGS is highly effective at removing snowflakes, rain streaks, and occlusions from camera lenses, resulting in visually appealing 3D scenes with high-quality rendered images in both synthetic and real-world scenarios.
For instance, in snowy scenes, WeatherGS effectively removes dense snowflakes that obstruct the view of the scene. In contrast, NeRF-based methods exhibit significant blurring due to over-smoothing of inconsistent views.
Similarly, in rainy scenes, WeatherGS successfully removes the 
raindrops adhering to the camera lens, whereas other methods only partially mitigate the weather effects. This effectiveness is attributed to the proposed dense-to-spare strategy, where the LED module easily extracts occlusions after the AEF module removes the dense weather particles.
Second,
although DerainNeRF can remove water droplets from the camera lens, it produces inconsistent reconstructions, leading to fragmented and blurred areas, which is also observed in vanilla-NeRF.
These inconsistencies stem from the varying appearances of snowflakes and rain streaks across different images, which disrupt NeRF model training. 

\noindent{\textbf{Quantitative Analysis}.}
The quantitative results of snowy and rainy scenes are listed in Table.~\ref{table: snowy} and Table.~\ref{table: rainy}.
The results show that our model achieves the highest PSNR, as well as the lowest LPIPS on average, indicating superior similarity to clean scenes compared to other approaches. However, in certain scenes, vanilla 3DGS and GS-W occasionally surpass our method in SSIM. This can be attributed to the use of preprocessed images in our model for training, whereas vanilla 3DGS and GS-W utilize the original input images. Although the latter can reconstruct weather particles, their sparse distribution in these cases does not significantly impact the SSIM metric.
Despite this, WeatherGS outperforms in reconstructing scenes that are closer to clean, realistic appearances across various weather conditions


\subsection{Ablation Studies}
\begin{figure}[ht]
    \centering
    \includegraphics[width=\linewidth]{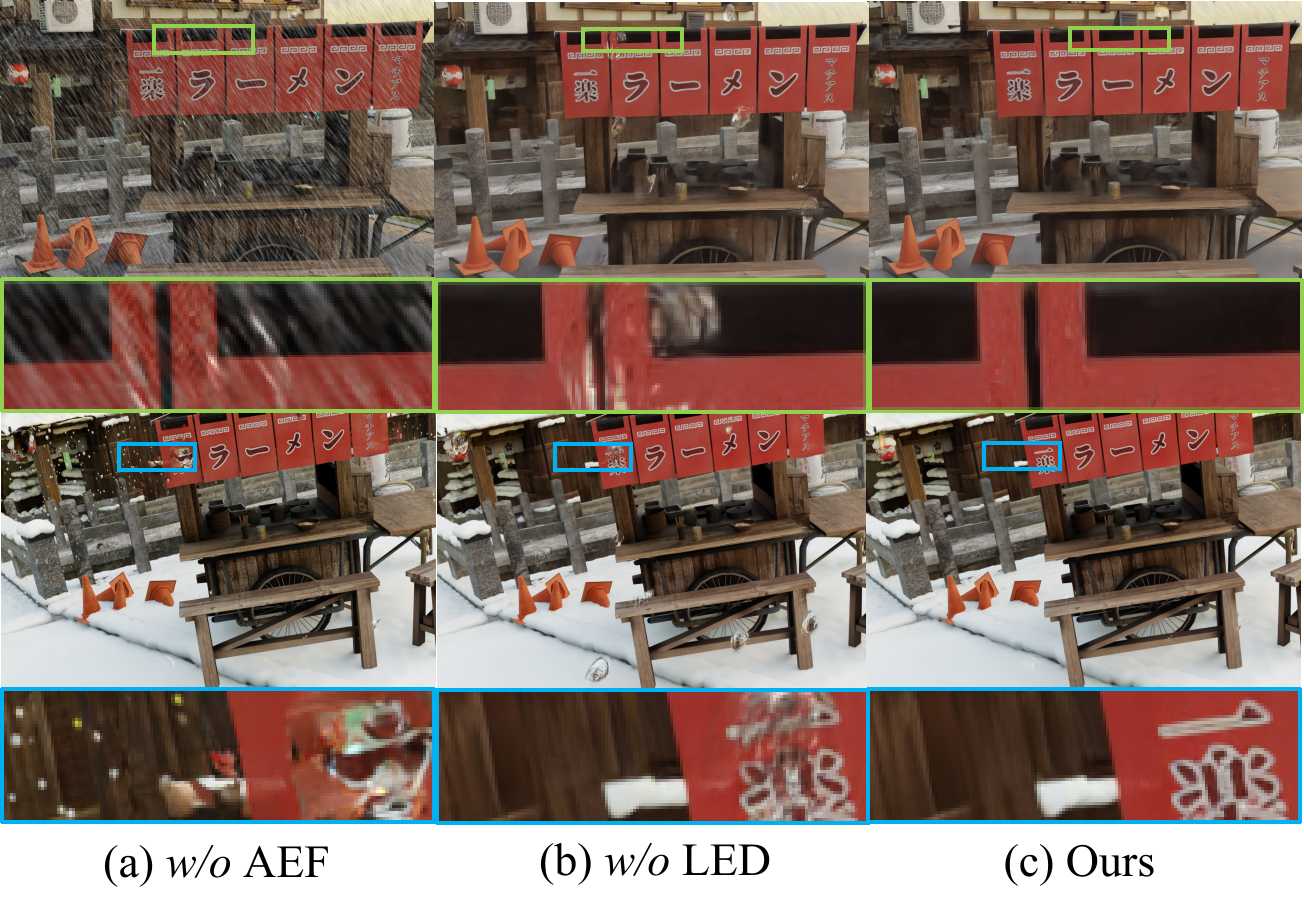}

    \caption{The comparison results of with and without AEF and LED.}
    \label{fig:ablation1}
    \vspace{-2mm}
\end{figure}

\begin{table}[h]
\centering
\caption{Comparison of PSNR, SSIM, and LPIPS metrics with and without AEF and LED. 
}

\begin{tabular}{lccc}
\toprule
\textbf{Model} & PSNR ↑ & SSIM ↑ & LPIPS ↓ \\
\midrule
w/o AEF & 24.3661 &\textbf{ 0.7452} & 0.2049 \\
w/o LED & 23.9460 & 0.7239 & 0.1986 \\
\textbf{Ours (Full)} & \textbf{24.6266} & 0.7395 & \textbf{0.1808} \\
\bottomrule
\end{tabular}
\label{tab:ablation1}

\vspace{-5mm}
\end{table}
To evaluate the effectiveness of our framework, we compare our model's performance with and without the preprocessing modules, as well as using different 3D reconstruction model backends. Comparisons are conducted across all synthetic scenes, including both snowy and rainy scenarios. 

\noindent \textbf{AEF and LED.} As illustrated in \cref{fig:ablation1}, incorporating the Atmospheric Effect Filter (AEF) alone effectively reduces weather particles in the processed images but fails to address occlusions on the camera lens. Similarly, using only the mask generated by the Lens Effect Detector (LED) to exclude occluded areas during training can remove occlusions but is insufficient for eliminating snowflakes and rain streaks. While integrating both components may introduce minor impacts to structural details due to the diffusion process, it significantly improves overall visual quality, as evidenced by the highest PSNR and lowest LPIPS shown in \cref{tab:ablation1}. 


\begin{figure}[ht]

    \centering
    \includegraphics[width=\linewidth]{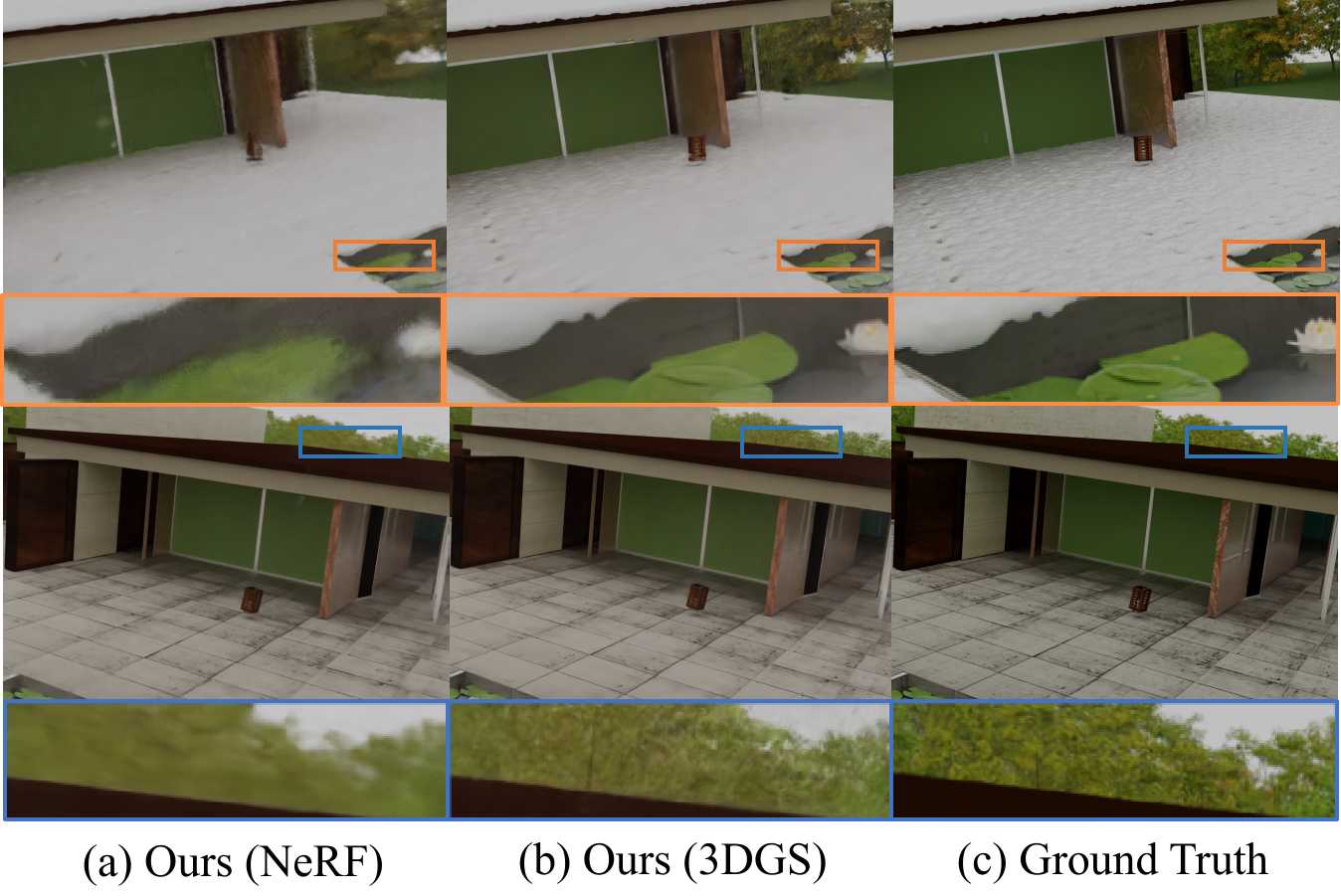}

    \caption{The comparison results of using NeRF and 3DGS backend.}
    \label{fig:scene_infer}
\end{figure}

\begin{table}[ht]
\centering
\scriptsize 
\setlength{\tabcolsep}{4pt} 
\renewcommand{\arraystretch}{1.2} 
\caption{Comparison of Average Performance Metrics, Training Time, and Rendering Time for Different Models}
\begin{tabular}{lcccccc}
\toprule
\multirow{2}{*}{\textbf{Model}} & \multicolumn{3}{c}{\textbf{Image Quality}} & \multicolumn{3}{c}{\textbf{Computational Performance}} \\ 
\cmidrule(lr){2-4} \cmidrule(lr){5-7} 
 & PSNR & SSIM  & LPIPS  & GPU Mem. & Training & Render \\
\midrule
Ours-NeRF  & 22.288 & 0.648 & 0.363 & 7.8G & $\sim$ 395 mins & 32.3 sec \\
\textbf{Ours-3DGS}  & \textbf{24.627} & \textbf{0.740} & \textbf{0.181} & \textbf{1.1G} & \textbf{$\sim$ 28 mins} & \textbf{0.02 sec} \\
\bottomrule
\end{tabular}
\label{Table:3d_computation_cost}
\vspace{-2mm}
\end{table}

\noindent \textbf{3D Reconstruction Model.} To highlight the advantages of utilizing 3DGS as the backend for our 3D reconstruction model, we perform a comparative analysis between our approach with NeRF and 3DGS as the underlying frameworks. 
We evaluate both image quality and computational performance including GPU memory usage, training time, and rendering time. 
As demonstrated in Fig.~\ref{fig:scene_infer} and Table.~\ref{Table:3d_computation_cost}, while our method using NeRF is capable of reconstructing relatively clear scenes, it exhibits notable blur effects and diminished texture details. Furthermore, the significantly higher GPU memory requirements, along with prolonged training and rendering times, greatly limit the practical applicability of the NeRF-based approach. In contrast, by employing 3DGS as the backend, our method achieves superior scene reconstruction with greater texture fidelity, while also demonstrating improved computational efficiency. This makes our approach more feasible for real-time applications and scalable to more complex environments.

\section{Discussion and Conclusion }
\label{sec:conclusion}
In this work, we introduce WeatherGS, a novel approach for 3D scene reconstruction from multi-view images affected by adverse weather, built on 3D Gaussian Splatting (3DGS). By utilizing a specially designed \textit{dense-to-sparse} mechanism, WeatherGS effectively addresses the challenges posed by weather-induced artifacts, such as dense particles and severe lens occlusions, significantly enhancing the generalization capability of 3DGS in outdoor environments. Furthermore, we establish a diverse and challenging benchmark to evaluate the performance of 3D reconstruction models under various weather conditions. Both qualitative and quantitative results on this benchmark demonstrate that WeatherGS consistently outperforms state-of-the-art reconstruction methods, rendering clear scenes regardless of weather conditions.




{\small
\bibliographystyle{IEEEtran}
\bibliography{references}
}

\end{document}